\documentclass{article}

\title{Cardiologist-Level Arrhythmia Detection with Convolutional Neural Networks}
\usepackage{times}
\usepackage{graphicx} 

\usepackage{booktabs}      

\usepackage{algorithm}
\usepackage{algorithmic}
\usepackage[caption = false]{subfig}
\usepackage{hyperref}
\usepackage{array}

\usepackage[accepted]{icml2015}

\begin{document} 

\twocolumn[
\icmltitle{Cardiologist-Level Arrhythmia Detection with Convolutional Neural Networks}

\icmlauthor{Pranav Rajpurkar$^*$}{pranavsr@cs.stanford.edu}
\icmlauthor{Awni Y. Hannun$^*$}{awni@cs.stanford.edu}
\icmlauthor{Masoumeh Haghpanahi}{mhaghpanahi@irhythmtech.com}
\icmlauthor{Codie Bourn}{cbourn@irhythmtech.com}
\icmlauthor{Andrew Y. Ng}{ang@cs.stanford.edu}

\renewcommand{\thefootnote}{\fnsymbol{footnote}}

\vskip 0.3in
]

\begin{abstract} 
We develop an algorithm which exceeds the performance of board certified cardiologists in detecting a wide range of heart arrhythmias from electrocardiograms recorded with a single-lead wearable monitor. We build a dataset with more than 500 times the number of unique patients than previously studied corpora. On this dataset, we train a 34-layer convolutional neural network which maps a sequence of ECG samples to a sequence of rhythm classes. Committees of board-certified cardiologists annotate a gold standard test set on which we compare the performance of our model to that of 6 other individual cardiologists. We exceed the average cardiologist performance in both recall (sensitivity) and precision (positive predictive value).
\end{abstract}

\section{Introduction}
{ \let\thefootnote\relax\footnote{$^*$Authors contributed equally.}}
{\let\thefootnote\relax\footnote{Project website at \url{https://stanfordmlgroup.github.io/projects/ecg}}}

\begin{figure}[ht]
  \centering
  \includegraphics[width=\linewidth]{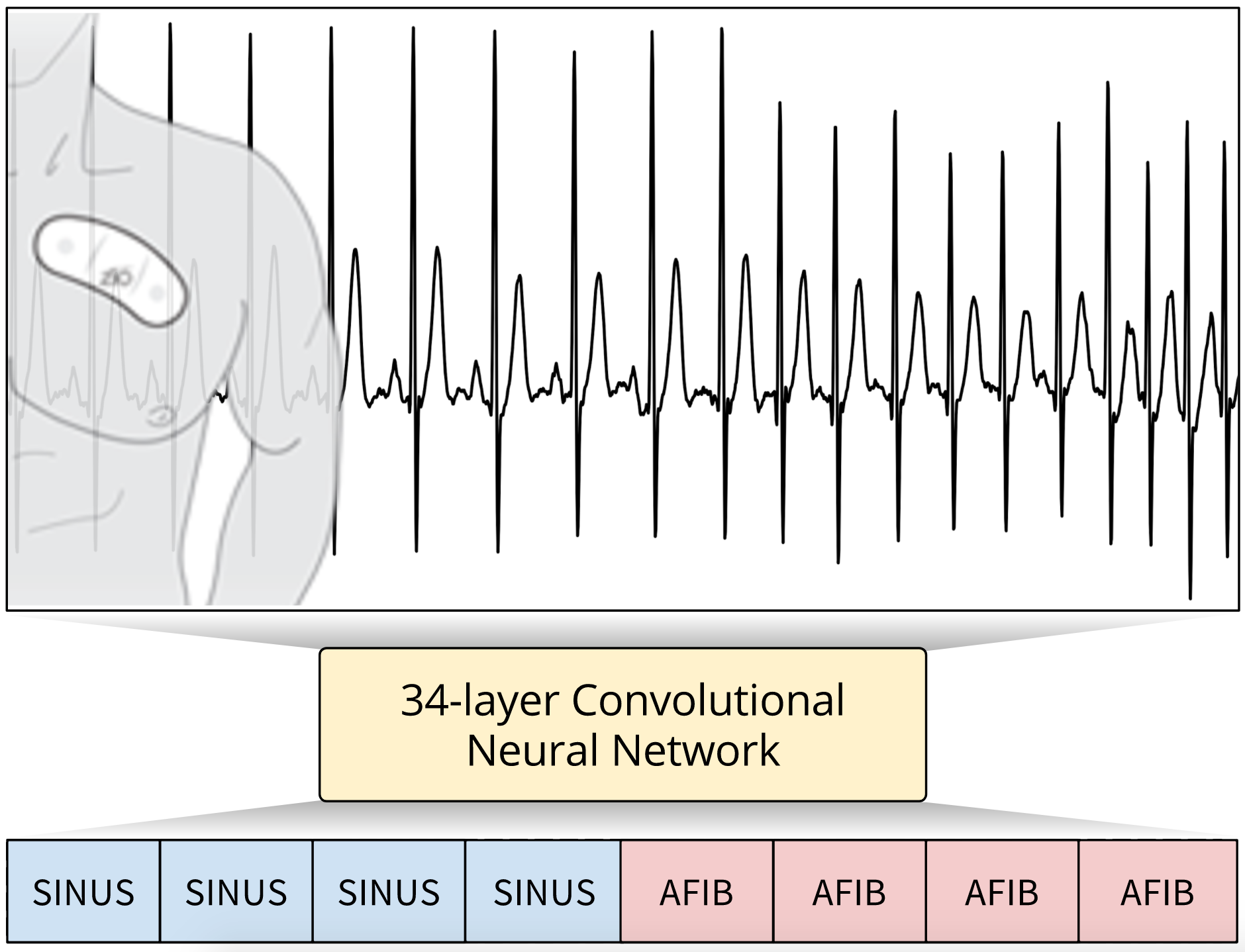}
  \caption{
    Our trained convolutional neural network correctly detecting the sinus rhythm (SINUS) and Atrial Fibrillation (AFIB) from this ECG recorded with a single-lead wearable heart monitor.
  }
  \label{fig:record}
\end{figure}
We develop a model which can diagnose irregular heart rhythms, also known as arrhythmias, from single-lead ECG signals better than a cardiologist. Key to exceeding expert performance is a deep convolutional network which can map a sequence of ECG samples to a sequence of arrhythmia annotations along with a novel dataset two orders of magnitude larger than previous datasets of its kind.

Many heart diseases, including Myocardial Infarction, AV Block, Ventricular Tachycardia and Atrial Fibrillation can all be diagnosed from ECG signals with an estimated 300 million ECGs recorded annually \cite{heden1996detection}. We investigate the task of arrhythmia detection from the ECG record. This is known to be a challenging task for computers but can usually be determined by an expert from a single, well-placed lead.

Arrhythmia detection from ECG recordings is usually performed by expert technicians and cardiologists given the high error rates of computerized interpretation.  One study found that of all the computer predictions for non-sinus rhythms, only about 50\% were correct \cite{shah2007errors}; in another study, only 1 out of every 7 presentations of second degree AV block were correctly recognized by the algorithm \cite{guglin2006common}. To automatically detect heart arrhythmias in an ECG, an algorithm must implicitly recognize the distinct wave types and discern the complex relationships between them over time. This is difficult due to the variability in wave morphology between patients as well as the presence of noise.

We train a 34-layer convolutional neural network (CNN) to detect arrhythmias in arbitrary length ECG time-series. Figure~\ref{fig:record} shows an example of an input to the model. In addition to classifying noise and the sinus rhythm, the network learns to classify and segment twelve arrhythmia types present in the time-series. The model is trained end-to-end on a single-lead ECG signal sampled at 200Hz and a sequence of annotations for every second of the ECG as supervision. To make the optimization of such a deep model tractable, we use residual connections and batch-normalization \cite{he2016deep, ioffe2015batch}. The depth increases both the non-linearity of the computation as well as the size of the context window for each classification decision.

We construct a dataset 500 times larger than other datasets of its kind \cite{moody2001impact, goldberger2000physiobank}. One of the most popular previous datasets, the MIT-BIH corpus contains ECG recordings from 47 unique patients. In contrast, we collect and annotate a dataset of about 30,000 unique patients from a pool of nearly 300,000 patients who have used the Zio Patch monitor\footnote[1]{iRhythm Technologies, San Francisco, California} \cite{turakhia2013diagnostic}. We intentionally select patients exhibiting abnormal rhythms in order to make the class balance of the dataset more even and thus the likelihood of observing unusual heart-activity high.

We test our model against board-certified cardiologists. A committee of three cardiologists serve as gold-standard annotators for the 336 examples in the test set. Our model exceeds the individual expert performance on both recall (sensitivity), and precision (positive predictive value) on this test set.

\section{Model}
\label{model}
\begin{figure}[ht!]
  \centering
  \includegraphics[height=15cm]{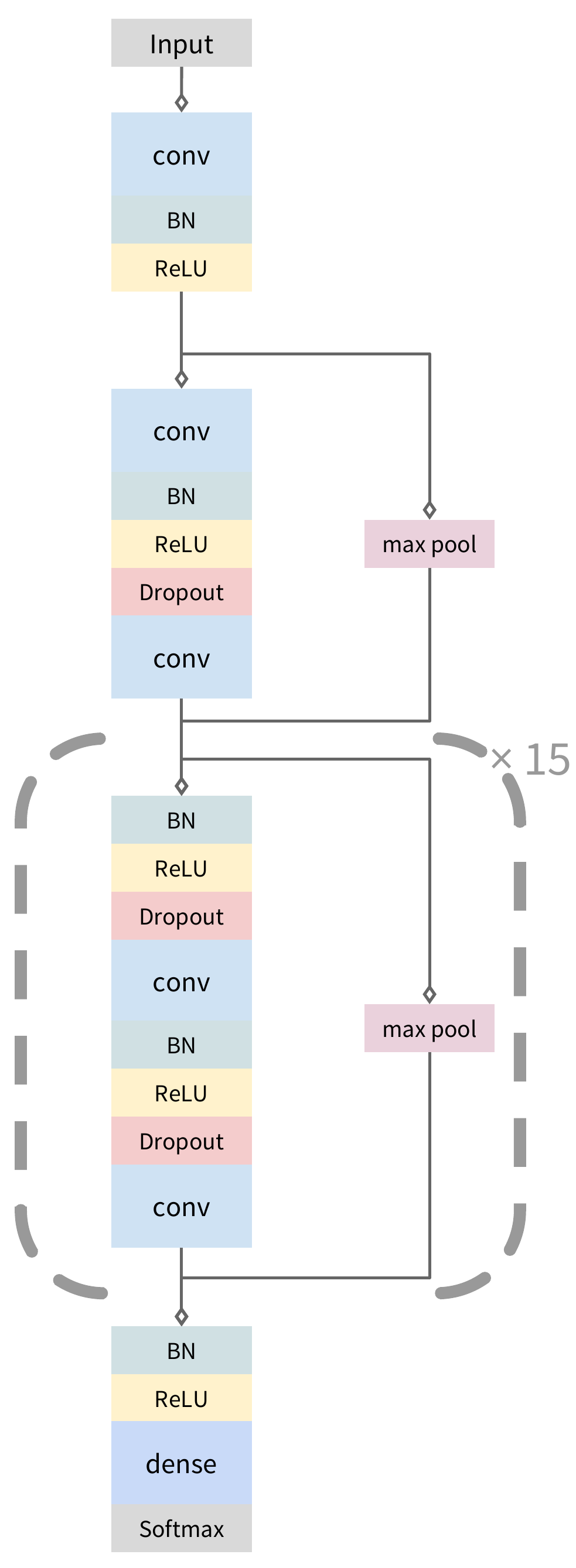}
  \caption{The architecture of the network. The first and last layer are special-cased due to the pre-activation residual blocks. Overall, the network contains 33 layers of convolution followed by a fully-connected layer and a softmax.}
  \label{fig:net}
\end{figure}

\subsection*{Problem Formulation}
The ECG arrhythmia detection task is a sequence-to-sequence task which takes as input an ECG signal $X=[x_1,.. x_k]$, and outputs a sequence of labels $r=[r_1, ... r_n],$ such that each $r_i$ can take on one of $m$ different rhythm classes. Each output label corresponds to a segment of the input. Together the output labels cover the full sequence.

For a single example in the training set, we optimize the cross-entropy objective function
\[
 \mathcal{L}(X, r) = \frac{1}{n} \sum_{i=1}^n \log p(R = r_i \mid X)
\]
where $p(\cdot)$ is the probability the network assigns to the $i$-th output taking on the value $r_i$.

\subsection*{Model Architecture and Training}
We use a convolutional neural network for the sequence-to-sequence learning task. The high-level architecture of the network is shown in Figure~\ref{fig:net}. The network takes as input a time-series of raw ECG signal, and outputs a sequence of label predictions. The 30 second long ECG signal is sampled at 200Hz, and the model outputs a new prediction once every second. We arrive at an architecture which is 33 layers of convolution followed by a fully connected layer and a softmax. 

In order to make the optimization of such a network tractable, we employ shortcut connections in a similar manner to those found in the Residual Network architecture \cite{DBLP:journals/corr/HeZRS15}. The shortcut connections between neural-network layers optimize training by allowing information to propagate well in very deep neural networks. Before the input is fed into the network, it is normalized using a robust normalization strategy. The network consists of $16$ residual blocks with $2$ convolutional layers per block. The convolutional layers all have a filter length of $16$ and have $64k$ filters, where $k$ starts out as $1$ and is incremented every $4$-th residual block. Every alternate residual block subsamples its inputs by a factor of $2$, thus the original input is ultimately subsampled by a factor of $2^8$. When a residual block subsamples the input, the corresponding shortcut connections also subsample their input using a Max Pooling operation with the same subsample factor. 

Before each convolutional layer we apply Batch Normalization \cite{ioffe2015batch} and a rectified linear activation, adopting the pre-activation block design \cite{DBLP:journals/corr/HeZR016}. The first and last layers of the network are special-cased due to this pre-activation block structure. We also apply Dropout \cite{srivastava2014dropout} between the convolutional layers and after the non-linearity. The final fully connected layer and softmax activation produce a distribution over the $14$ output classes for each time-step.

We train the networks from scratch, initializing the weights of the convolutional layers as in \cite{DBLP:journals/corr/HeZR015}. We use the Adam \cite{kingma2014adam} optimizer with the default parameters and reduce the learning rate by a factor of $10$ when the validation loss stops improving. We save the best model as evaluated on the validation set during the optimization process.
[ht]
\begin{figure}
  \centering
  \includegraphics[width=\linewidth]{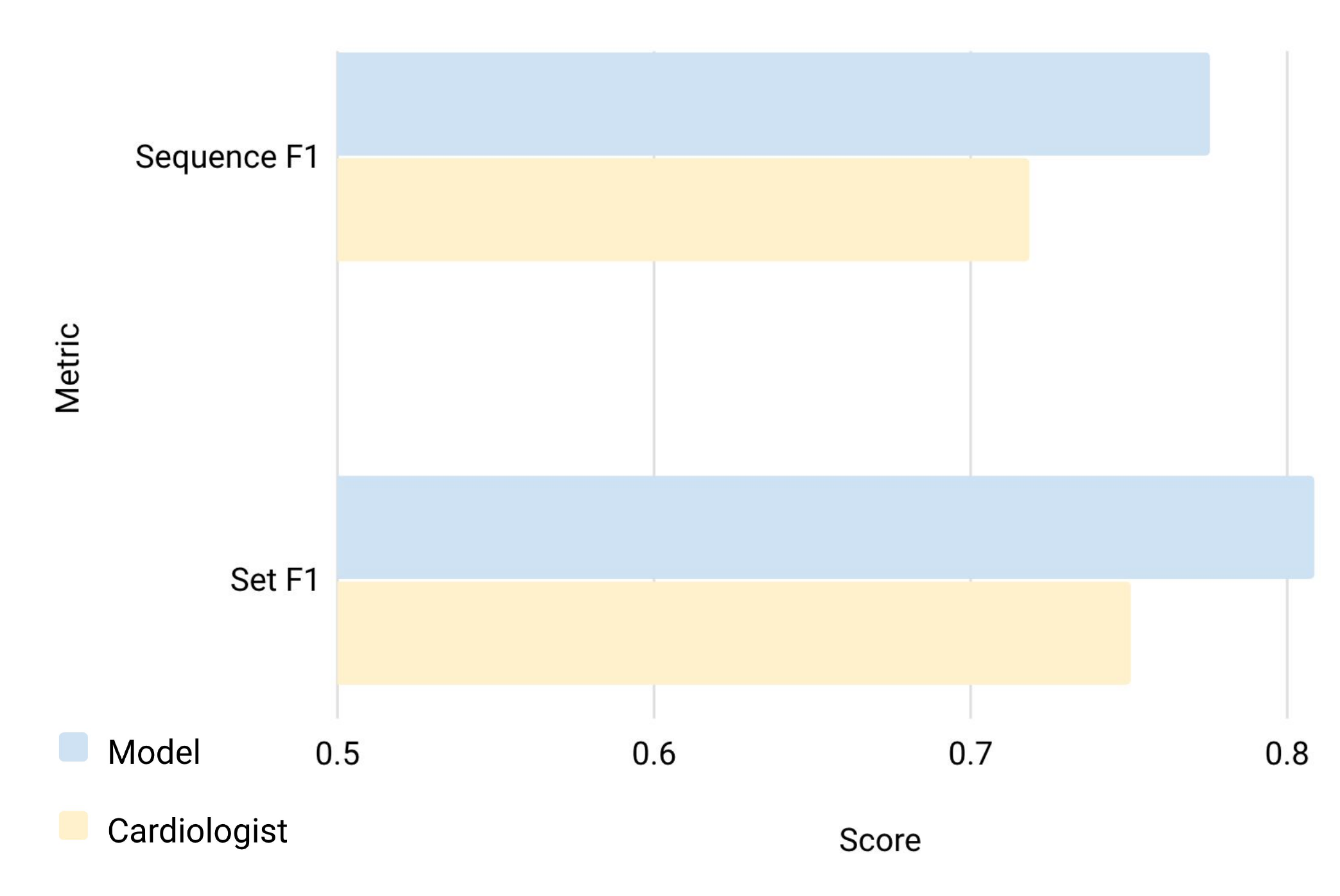}
  \caption{
    Evaluated on the test set, the model outperforms the average cardiologist score on both the Sequence and the Set F1 metrics.
  }
  \label{fig:model_vs_cardiol}
\end{figure}

\section{Data}
\label{data}
\subsection*{Training}
We collect and annotate a dataset of 64,121 ECG records from 29,163 patients. The ECG data is sampled at a frequency of 200 Hz and is collected from a single-lead, noninvasive and  continuous monitoring device called the Zio Patch which has a wear period up to 14 days \cite{turakhia2013diagnostic}. Each ECG record in the training set is 30 seconds long and can contain more than one rhythm type. Each record is annotated by a clinical ECG expert: the expert highlights segments of the signal and marks it as corresponding to one of the 14 rhythm classes.

The 30 second records were annotated using a web-based ECG annotation tool designed for this work. Label annotations were done by a group of Certified Cardiographic Technicians who have completed extensive training in arrhythmia detection and a cardiographic certification examination by Cardiovascular Credentialing International. The technicians were guided through the interface before they could annotate records. All rhythms present in a strip were labeled from their corresponding onset to offset, resulting in full segmentation of the input ECG data. To improve labeling consistency among different annotators, specific rules were devised regarding each rhythm transition.

We split the dataset into a training and validation set. The training set contains 90\% of the data. We split the dataset so that there is no patient overlap between the training and validation sets (as well as the test set described below).

\subsection*{Testing}

We collect a test set of 336 records from 328 unique patients. For the test set, ground truth annotations for each record were obtained by a committee of three board-certified cardiologists; there are three committees responsible for different splits of the test set. The cardiologists discussed each individual record as a group and came to a consensus labeling. For each record in the test set we also collect 6 individual annotations from cardiologists not participating in the group. This is used to assess performance of the model compared to an individual cardiologist.

\subsection*{Rhythm Classes}
We identify 12 heart arrhythmias, sinus rhythm and noise for a total of 14 output classes. The arrhythmias are characterized by a variety of features. Table~\ref{tab:rhythms} in the Appendix shows an example of each rhythm type we classify. The noise label is assigned when the device is disconnected from the skin or when the baseline noise in the ECG makes identification of the underlying rhythm impossible.

The morphology of the ECG during a single heart-beat as well as the pattern of the activity of the heart over time determine the underlying rhythm. In some cases the distinction between the rhythms can be subtle yet critical for treatment. For example two forms of second degree AV Block, Mobitz I (Wenckebach) and Mobitz II (here referred to as AVB\_TYPE2) can be difficult to distinguish. Wenckebach is considered benign and Mobitz II is considered pathological, requiring immediate attention \cite{dubin1996rapid}. 

Table~\ref{tab:rhythms} in the Appendix also shows the number of unique patients in the training (including validation) set and test set for each rhythm type.

\begin{table}    
\centering
\begin{footnotesize}
\begin{tabular}{l c c c c}
\toprule
 & \multicolumn{2}{c}{Seq} & \multicolumn{2}{c}{Set} \\
\cmidrule{2-5}
 & Model & Cardiol. & Model & Cardiol. \\
\midrule
Class-level F1 Score & & & & \\
\midrule
 AFIB & \textbf{0.604} & 0.515 & \textbf{0.667} & 0.544 \\
 AFL & \textbf{0.687} & 0.635 & \textbf{0.679} & 0.646 \\
 AVB\_TYPE2 & \textbf{0.689} & 0.535 & \textbf{0.656} & 0.529 \\
 BIGEMINY & \textbf{0.897} & 0.837 & \textbf{0.870} & 0.849 \\
 CHB & \textbf{0.843} & 0.701 & \textbf{0.852} & 0.685 \\
 EAR & \textbf{0.519} & 0.476 & \textbf{0.571} & 0.529 \\
 IVR & \textbf{0.761} & 0.632 & \textbf{0.774} & 0.720 \\
 JUNCTIONAL & 0.670 & \textbf{0.684} & \textbf{0.783} & 0.674 \\
 NOISE & \textbf{0.823} & 0.768 & \textbf{0.704} & 0.689 \\
 SINUS & \textbf{0.879} & 0.847 & \textbf{0.939} & 0.907 \\
 SVT & \textbf{0.477} & 0.449 & \textbf{0.658} & 0.556 \\
 TRIGEMINY & \textbf{0.908} & 0.843 & \textbf{0.870} & 0.816 \\
 VT & 0.506 & \textbf{0.566} & 0.694 & \textbf{0.769} \\
 WENCKEBACH & \textbf{0.709} & 0.593 & \textbf{0.806} & 0.736 \\
\midrule\midrule
 Aggregate Results & & & & \\
\midrule
 Precision (PPV) & \textbf{0.800} & 0.723 & \textbf{0.809} & 0.763 \\
 Recall (Sensitivity) & \textbf{0.784} & 0.724 & \textbf{0.827} & 0.744 \\
 F1 & \textbf{0.776} & 0.719 & \textbf{0.809} & 0.751 \\
\bottomrule
\end{tabular}
\end{footnotesize}
\caption{The top part of the table gives a class-level comparison of the expert to the model F1 score for both the Sequence and the Set metrics. The bottom part of the table shows aggregate results over the full test set for precision, recall and F1 for both the Sequence and Set metrics.}
\label{tab:HumanVsModel}
\end{table}
\section{Results}
\label{results}
\subsection*{Evaluation Metrics}
We use two metrics to measure model accuracy, using the cardiologist committee annotations as the ground truth.

\textbf{Sequence Level Accuracy (F1):} We measure the average overlap between the prediction and the ground truth sequence labels. For every record, a model is required to make a prediction approximately once per second (every $256$ samples). These predictions are compared against the ground truth annotation.

\textbf{Set Level Accuracy (F1):} Instead of treating the labels for a record as a sequence, we consider the set of unique arrhythmias present in each $30$ second record as the ground truth annotation. Set Level Accuracy, unlike Sequence Level Accuracy, does not penalize for time-misalignment within a record. We report the F1 score between the unique class labels from the ground truth and those from the model prediction.

In both the Sequence and the Set case, we compute the F1 score for each class separately. We then compute the overall F1 (and precision and recall) as the class-frequency weighted mean.

\subsection*{Model vs. Cardiologist Performance}
We assess the cardiologist performance on the test set. Recall that each of the records in the test set has a ground truth label from a committee of three cardiologists as well as individual labels from a disjoint set of 6 other cardiologists. To assess cardiologist performance for each class, we take the average of all the individual cardiologist F1 scores using the group label as the ground truth annotation.

Table~\ref{tab:HumanVsModel} shows the breakdown of both cardiologist and model scores across the different rhythm classes. The model outperforms the average cardiologist performance on most rhythms, noticeably outperforming the cardiologists in the AV Block set of arrhythmias which includes Mobitz I (Wenckebach), Mobitz II (AVB\_Type2) and complete heart block (CHB). This is especially useful given the severity of Mobitz II and complete heart block and the importance of distinguishing these two from Wenckebach which is usually considered benign.

Table~\ref{tab:HumanVsModel} also compares the aggregate precision, recall and F1 for both  model and cardiologist compared to the ground truth annotations. The aggregate scores for the cardiologist are computed by taking the mean of the individual cardiologist scores. The model outperforms the cardiologist average in both precision and recall.

\section{Analysis}
\label{analysis}
\begin{figure}[t]
  \centering
  \includegraphics[width=\linewidth]{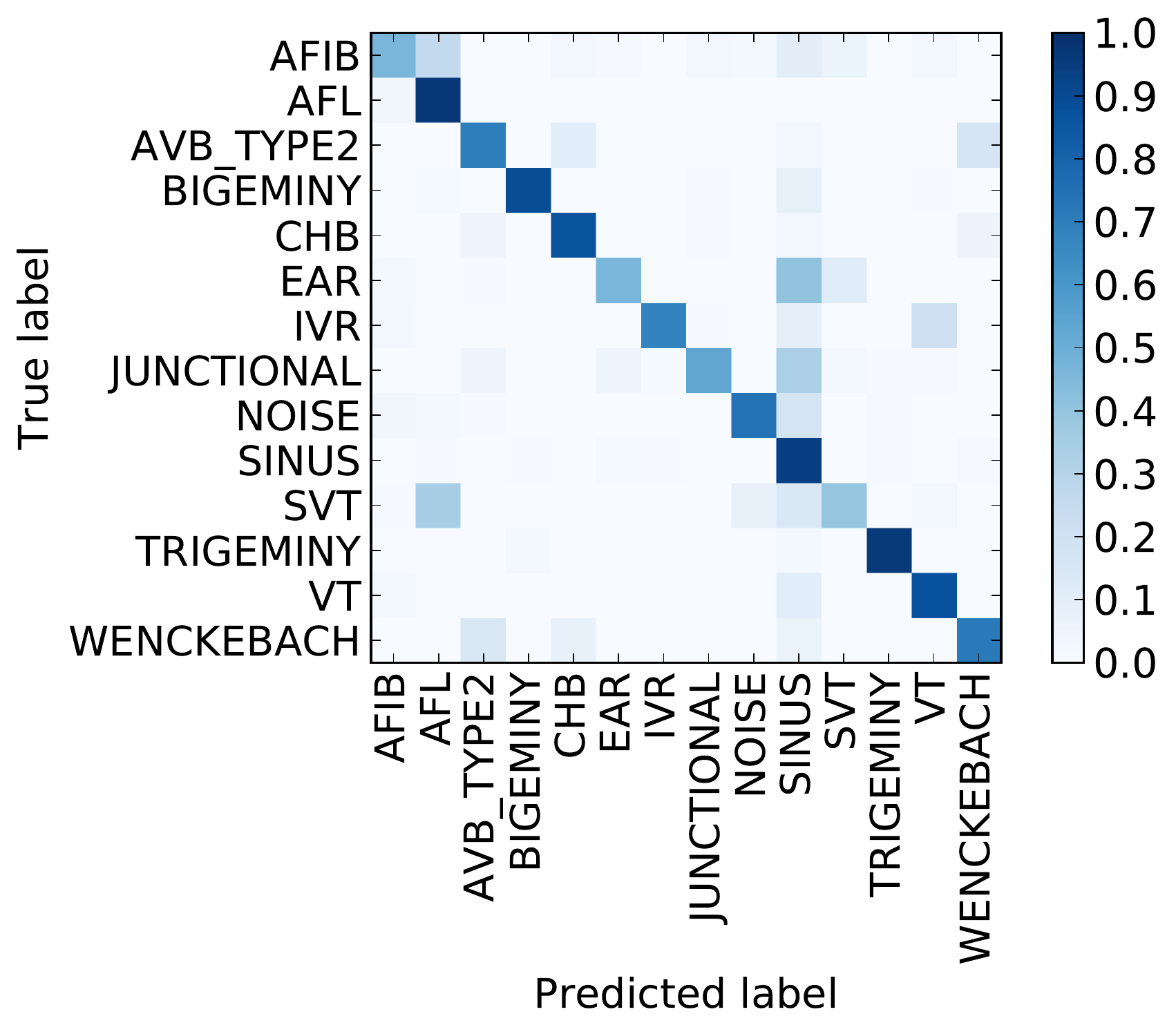}
  \caption{
    A confusion matrix for the model predictions on the test set. Many of the mistakes the model makes are not surprising. For example, confusing second degree AV Block (Type 2) with Wenckebach makes sense given the often similar expression of the two arrhythmias in the ECG record.
  }
  \label{fig:confusion}
\end{figure}

The model outperforms the average cardiologist score on both the sequence and the set F1 metrics. Figure~\ref{fig:confusion} shows a confusion matrix of the model predictions on the test set. Many arrhythmias are confused with the sinus rhythm. We expect that part of this is due to the sometimes ambiguous location of the exact onset and offset of the arrhythmia in the ECG record.

Often the mistakes made by the model are understandable. For example, confusing Wenckebach and AVB\_Type2 makes sense given that the two rhythms in general have very similar ECG morphologies. Similarly, Supraventricular Tachycardia (SVT) and Atrial Fibrillation (AFIB) are often confused with Atrial Flutter (AFL) which is understandable given that they are all atrial arrhythmias. We also note that Idioventricular Rhythm (IVR) is sometimes mistaken as Ventricular Tachycardia (VT), which again makes sense given that the two only differ in heart-rate and are difficult to distinguish close to the $100$ beats per minute delineation.

One of the most common confusions is between Ectopic Atrial Rhythm (EAR) and sinus rhythm. The main distinguishing criteria for this rhythm is an irregular P wave. This can be subtle to detect especially when the P wave has a small amplitude or when noise is present in the signal.

\section{Related Work}
\label{related}
Automatic high-accuracy methods for R-peak extraction have existed at least since the mid 1980's \cite{pan1985real}. Current algorithms for R-peak extraction tend to use wavelet transformations to compute features from the raw ECG followed by finely-tuned threshold based classifiers \cite{li1995detection, martinez2004wavelet}. Because accurate estimates of heart rate and heart rate variability can be extracted from R-peak features, feature-engineered algorithms are often used for coarse-grained heart rhythm classification, including detecting tachycardias (fast heart rate), bradycardias (slow heart rate), and irregular rhythms. However, such features alone are not sufficient to distinguish between most heart arrhythmias since features based on the atrial activity of the heart as well as other features pertaining to the QRS morphology are needed.

Much work has been done to automate the extraction of other features from the ECG. For example, beat classification is a common sub-problem of heart-arrhythmia classification. Drawing inspiration from automatic speech recognition, Hidden Markov models with Gaussian observation probability distributions have been applied to the task of beat detection \cite{coast1990approach}. Artificial neural networks have also been used for the task of beat detection \cite{melo2000arrhythmia}. While these models have achieved high-accuracy for some beat types, they are not yet sufficient for high-accuracy heart arrhythmia classification and segmentation. For example, \cite{artis1991detection} train a neural network to distinguish between Atrial Fibrillation and Sinus Rhythm on the MIT-BIH dataset. While the network can distinguish between these two classes with high-accuracy, it does not generalize to noisier single-lead recordings or classify among the full range of $15$ rhythms available in MIT-BIH. This is in part due to insufficient training data, and because the model also discards critical information in the feature extraction stage.

The most common dataset used to design and evaluate ECG algorithms is the MIT-BIH arrhythmia database \cite{moody2001impact} which consists of 48 half-hour strips of ECG data. Other commonly used datasets include the MIT-BIH Atrial Fibrillation dataset \cite{moody1983new} and the QT dataset \cite{laguna1997database}. While useful benchmarks for R-peak extraction and beat-level annotations, these datasets are too small for fine-grained arrhythmia classification. The number of unique patients is in the single digit hundreds or fewer for these benchmarks. A recently released dataset captured from the AliveCor ECG monitor contains about 7000 records \cite{clifford2017}. These records only have annotations for Atrial Fibrillation; all other arrhythmias are grouped into a single bucket. The dataset we develop contains 29,163 unique patients and $14$ classes with hundreds of unique examples for the rarest arrhythmias.

Machine learning models based on deep neural networks have consistently been able to approach and often exceed human agreement rates when large annotated datasets are available \cite{amodei2016deep, xiong2016achieving,he2015delving}. These approaches have also proven to be effective in healthcare applications, particularly in medical imaging where pretrained ImageNet models can be applied \cite{esteva2017dermatologist, gulshan2016development}. We draw on work in automatic speech recognition for processing time-series with deep convolutional neural networks and recurrent neural networks \cite{Hannun2014deepspeech, sainath2013deep}, and techniques in deep learning to make the optimization of these models tractable \cite{he2016deep, he2016identity, ioffe2015batch}.

\section{Conclusion}
\label{conclusion}
We develop a model which exceeds the cardiologist performance in detecting a wide range of heart arrhythmias from single-lead ECG records. Key to the performance of the model is a large annotated dataset and a very deep convolutional network which can map a sequence of ECG samples to a sequence of arrhythmia annotations. 

On the clinical side, future work should investigate extending the set of arrhythmias and other forms of heart disease which can be automatically detected with high-accuracy from single or multiple lead ECG records. For example we do not detect Ventricular Flutter or Fibrillation. We also do not detect Left or Right Ventricular Hypertrophy, Myocardial Infarction or a number of other heart diseases which do not necessarily exhibit as arrhythmias. Some of these may be difficult or even impossible to detect on a single-lead ECG but can often be seen on a multiple-lead ECG.

Given that more than 300 million ECGs are recorded annually, high-accuracy diagnosis from ECG can save expert clinicians and cardiologists considerable time and decrease the number of misdiagnoses. Furthermore, we hope that this technology coupled with low-cost ECG devices enables more widespread use of the ECG as a diagnostic tool in places where access to a cardiologist is difficult.

\section*{Acknowledgements}
We thank Geoffrey H. Tison MD, MPH of UCSF for helpful feedback on the experiments and references.

\bibliographystyle{icml2015}
\bibliography{refs}

\begin{table*}
\vspace{-12em}
\section*{Appendix}
\vspace{4em}
\centering
\small
\begin{tabular}{m{2cm} m{2cm} ccc}
\toprule
 & &  & Train + Val & Test \\
Class & Description & Example & Patients & Patients  \\
\midrule
	AFIB & Atrial Fibrillation & \parbox[c]{13em}{\vspace*{1mm}\includegraphics[height=8em]{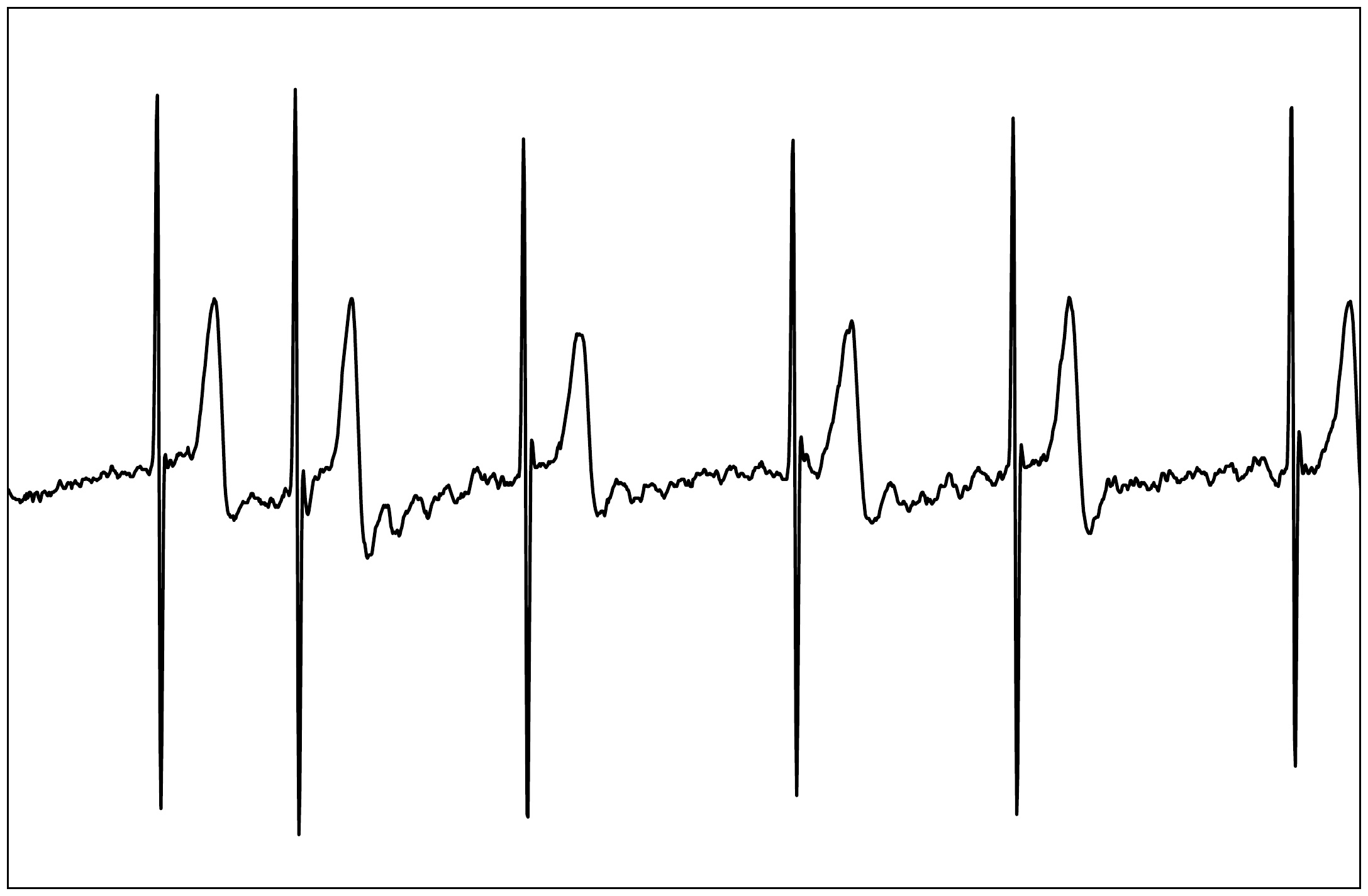}} & 4638 & 44 \\
    AFL & Atrial Flutter & \parbox[c]{13em}{\vspace*{1mm}\includegraphics[height=8em]{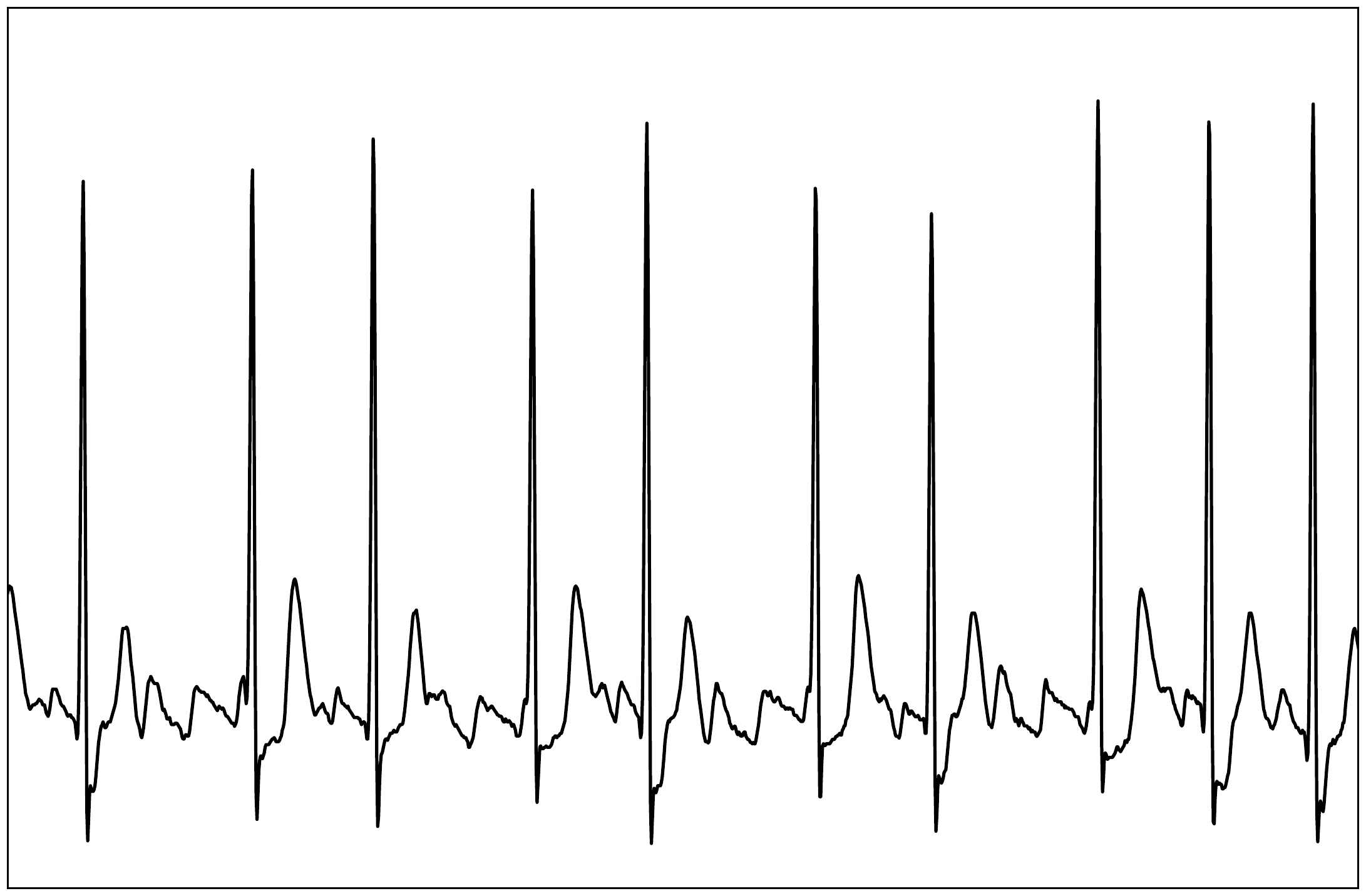}} & 3805 & 20 \\
    AVB\_TYPE2 & Second degree AV Block Type 2 (Mobitz II) & \parbox[c]{13em}{\includegraphics[height=8em]{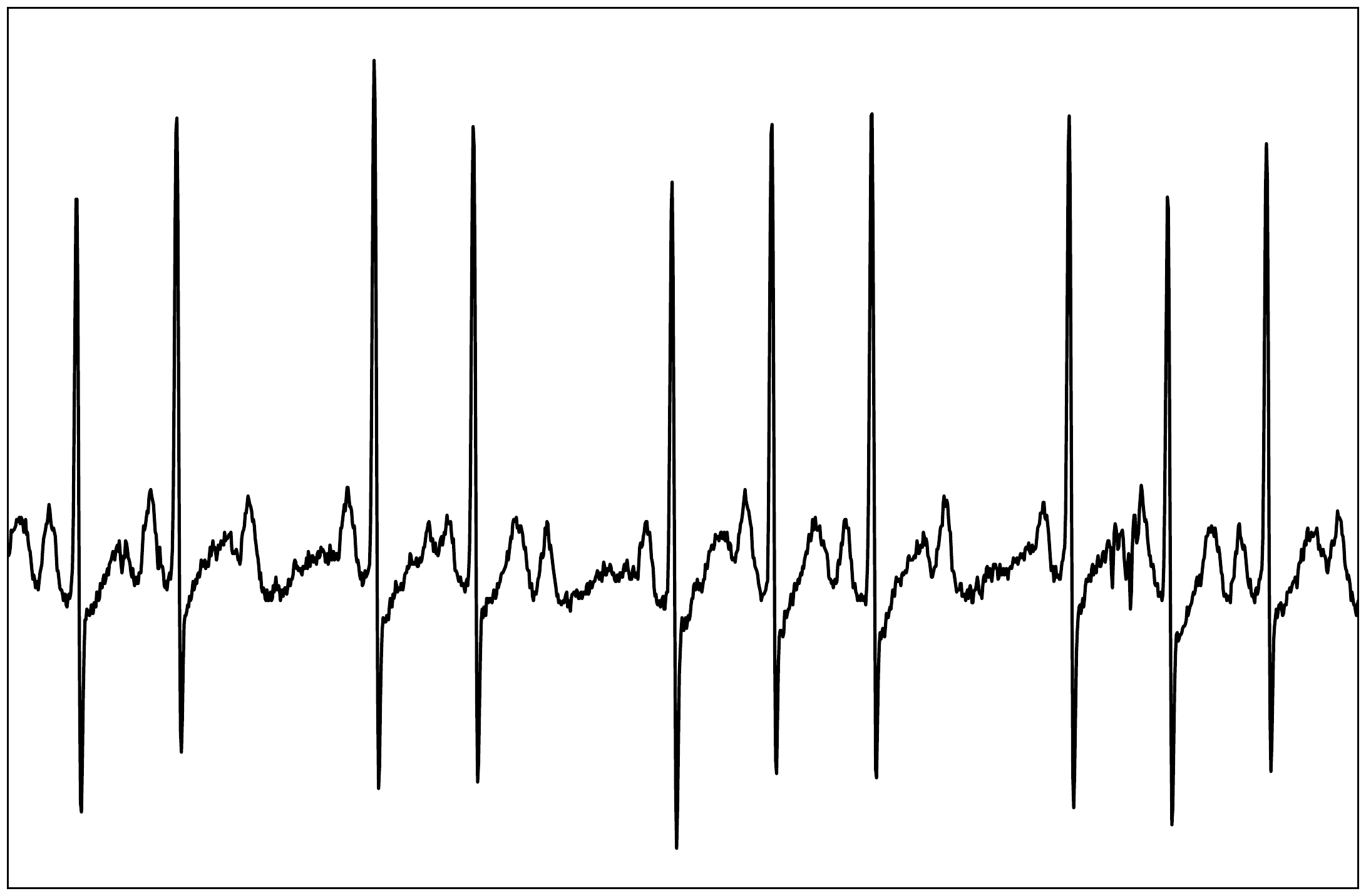}} & 1905  & 28 \\
    BIGEMINY & Ventricular Bigeminy & \parbox[c]{13em}{\includegraphics[height=8em]{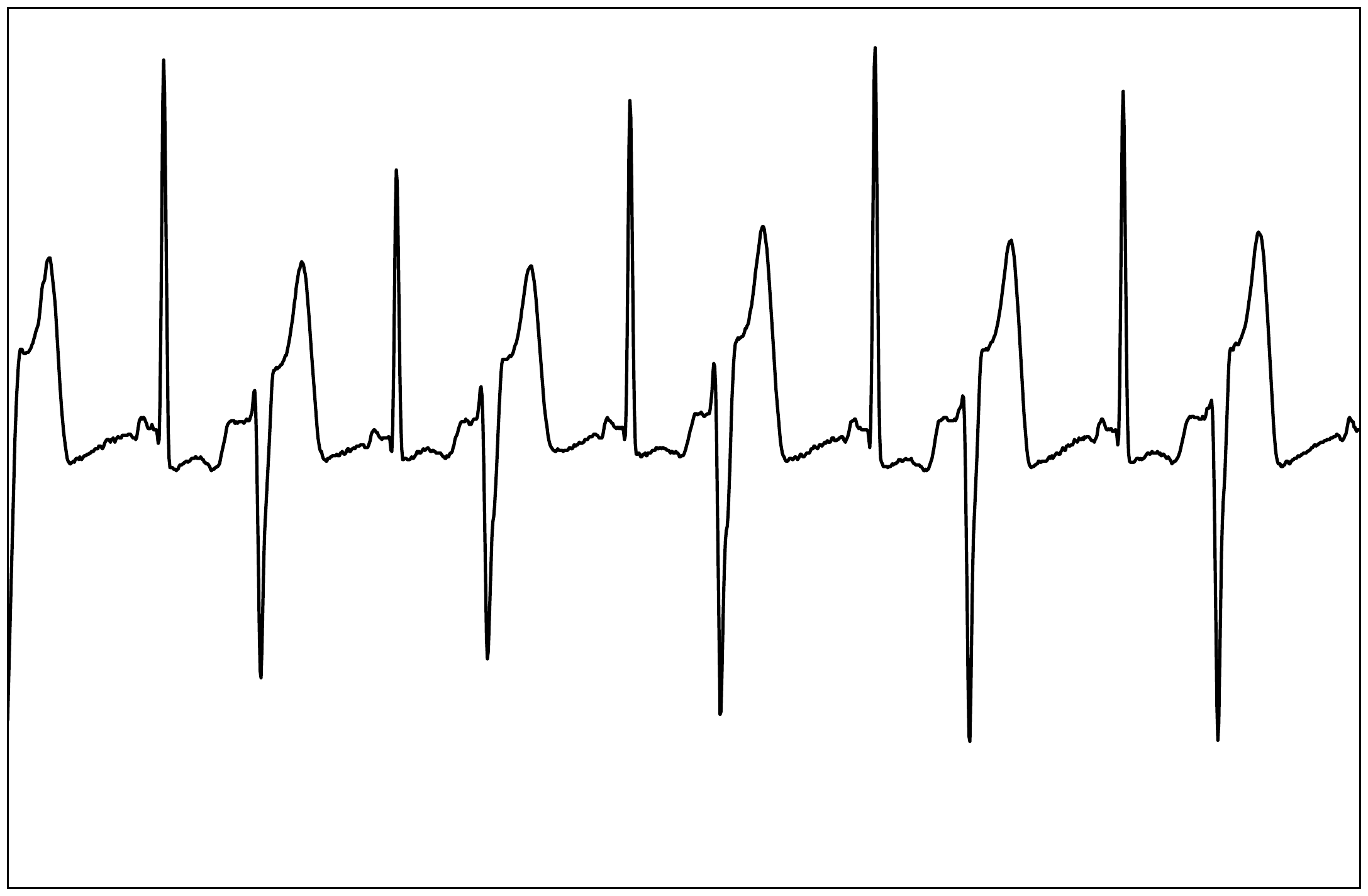}} & 2855 & 22 \\
    CHB & Complete Heart Block & \parbox[c]{13em}{\includegraphics[height=8em]{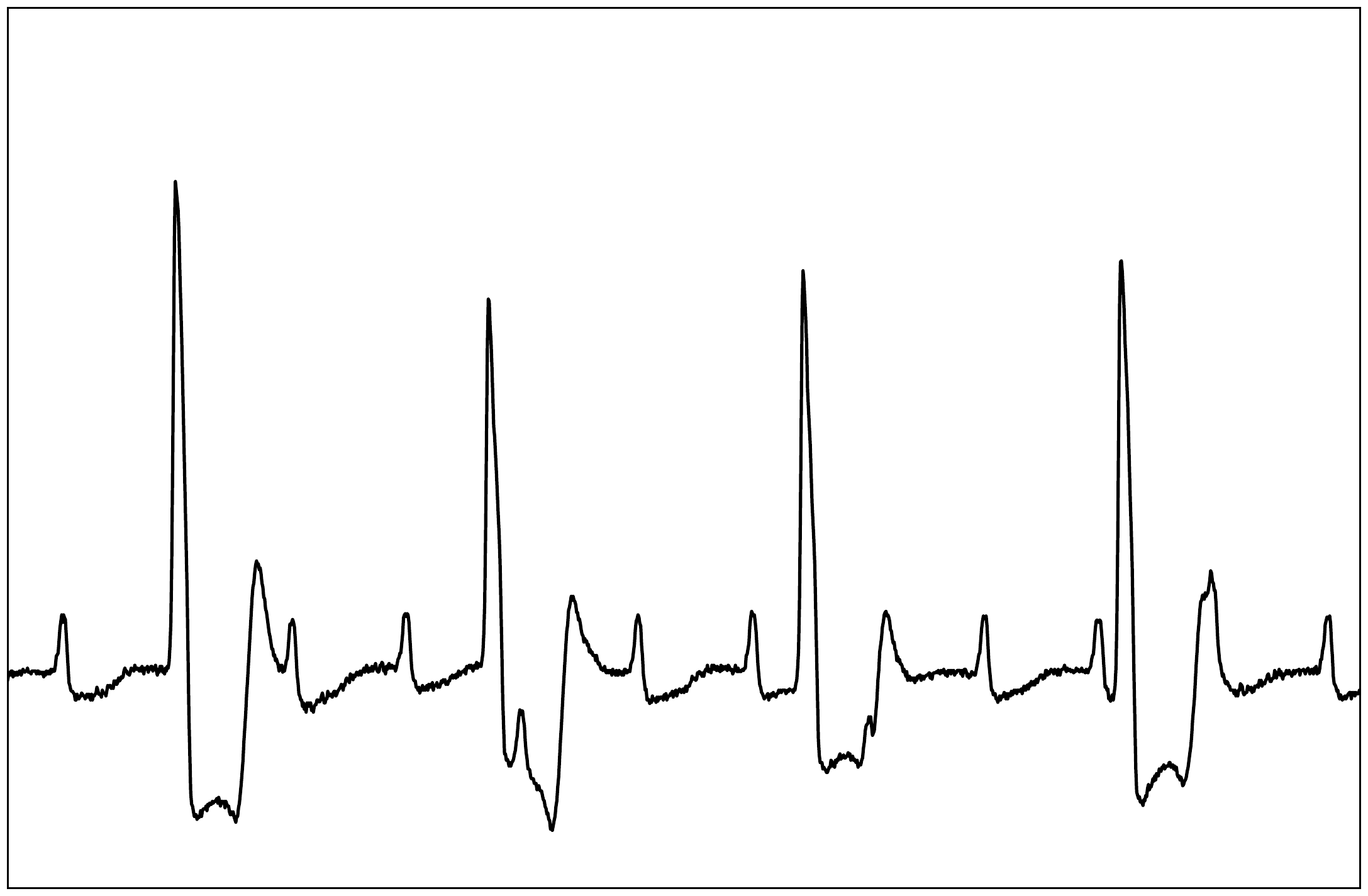}} & 843 & 26 \\
    EAR & Ectopic Atrial Rhythm & \parbox[c]{13em}{\includegraphics[height=8em]{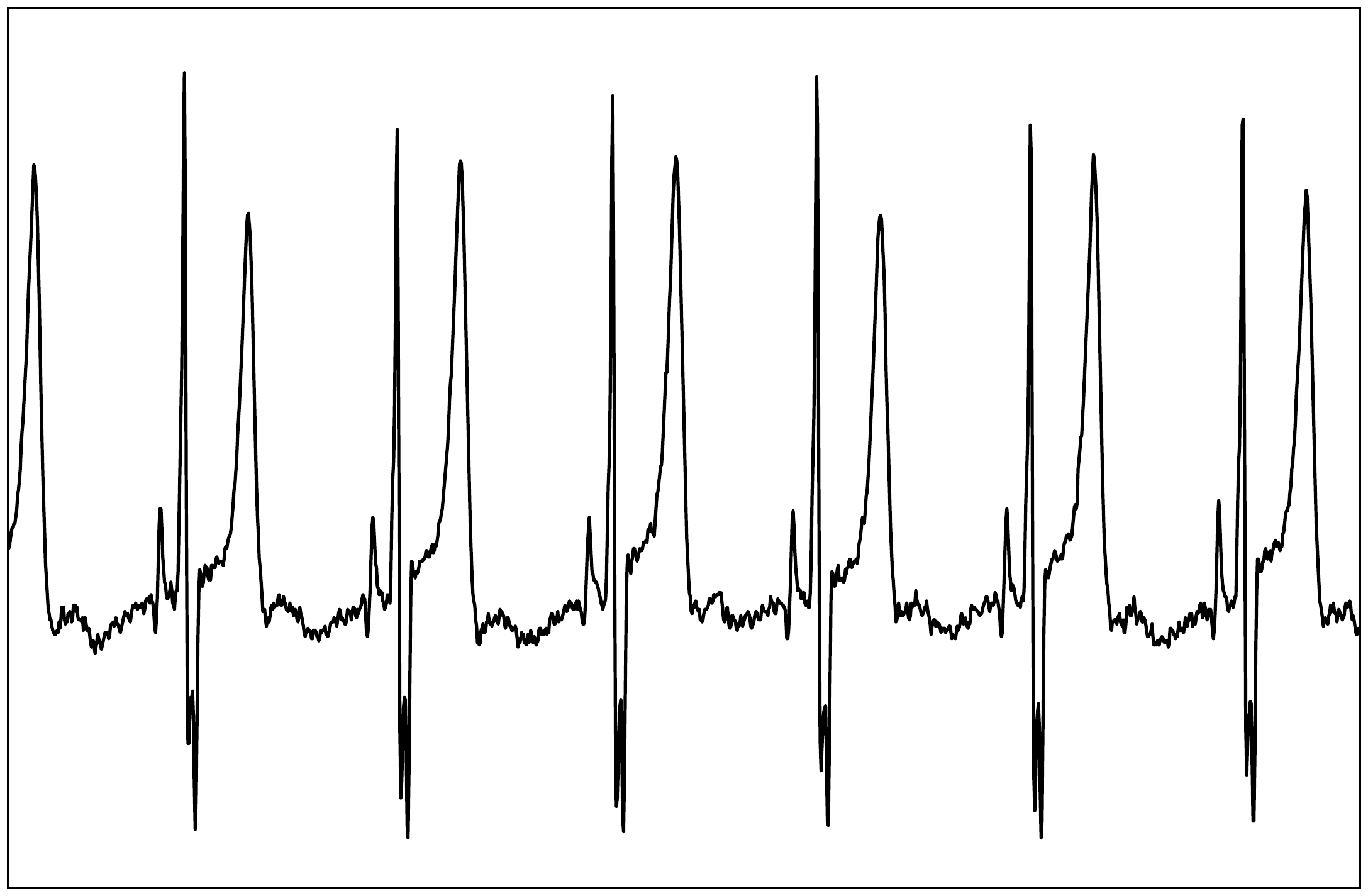}} & 2623 & 22 \\
    IVR & Idioventricular Rhythm & \parbox[c]{13em}{\includegraphics[height=8em]{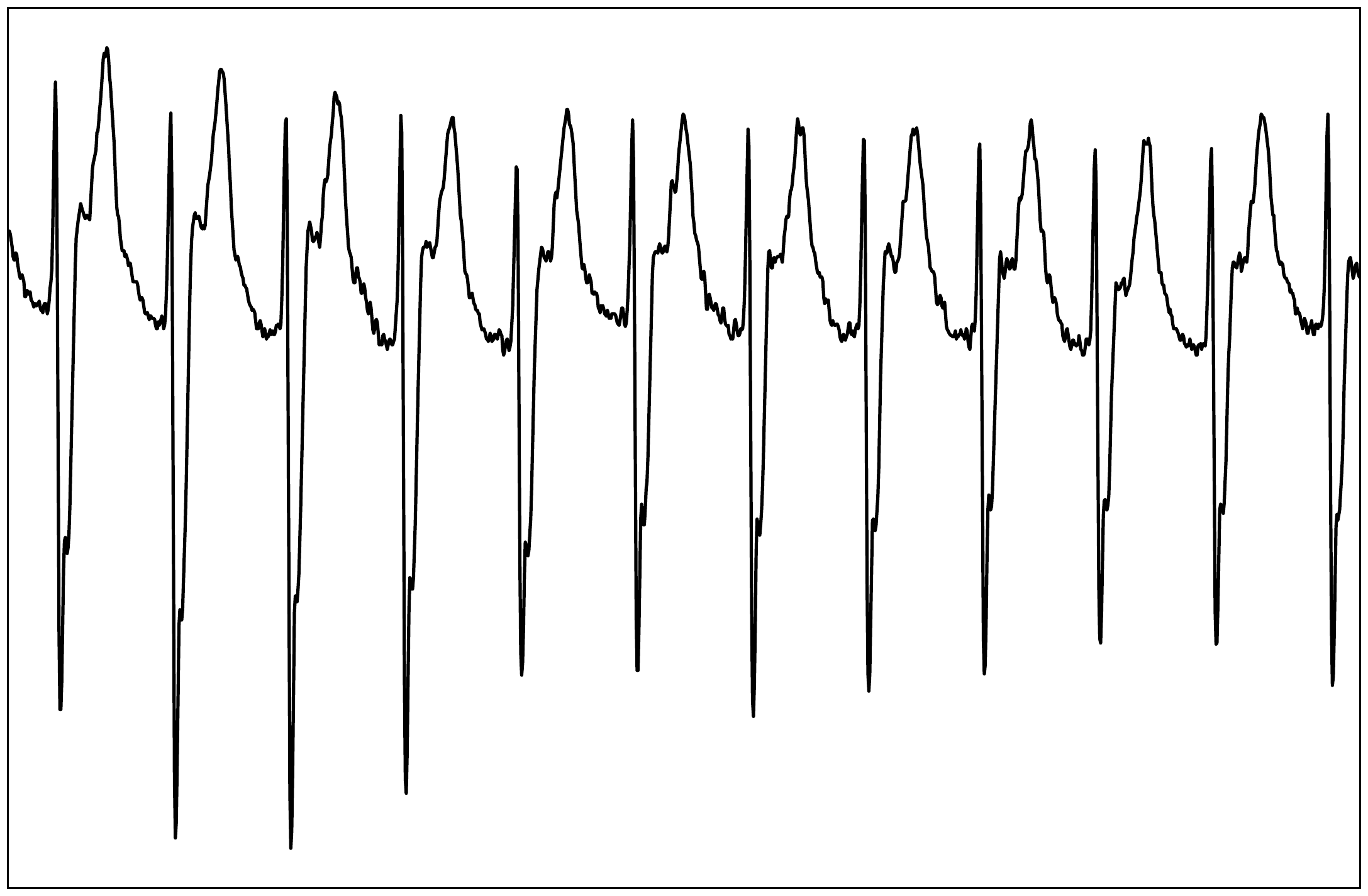}} & 1962 & 34 \\
\end{tabular}
\end{table*}

\begin{table*}
\centering
\small
\begin{tabular}{m{2cm} m{2cm} ccc}
\toprule
 & &  & Train + Val & Test \\
Class & Description & Example & Patients & Patients  \\
\midrule
	JUNCTIONAL & Junctional Rhythm & \parbox[c]{13em}{\vspace*{1mm}\includegraphics[height=8em]{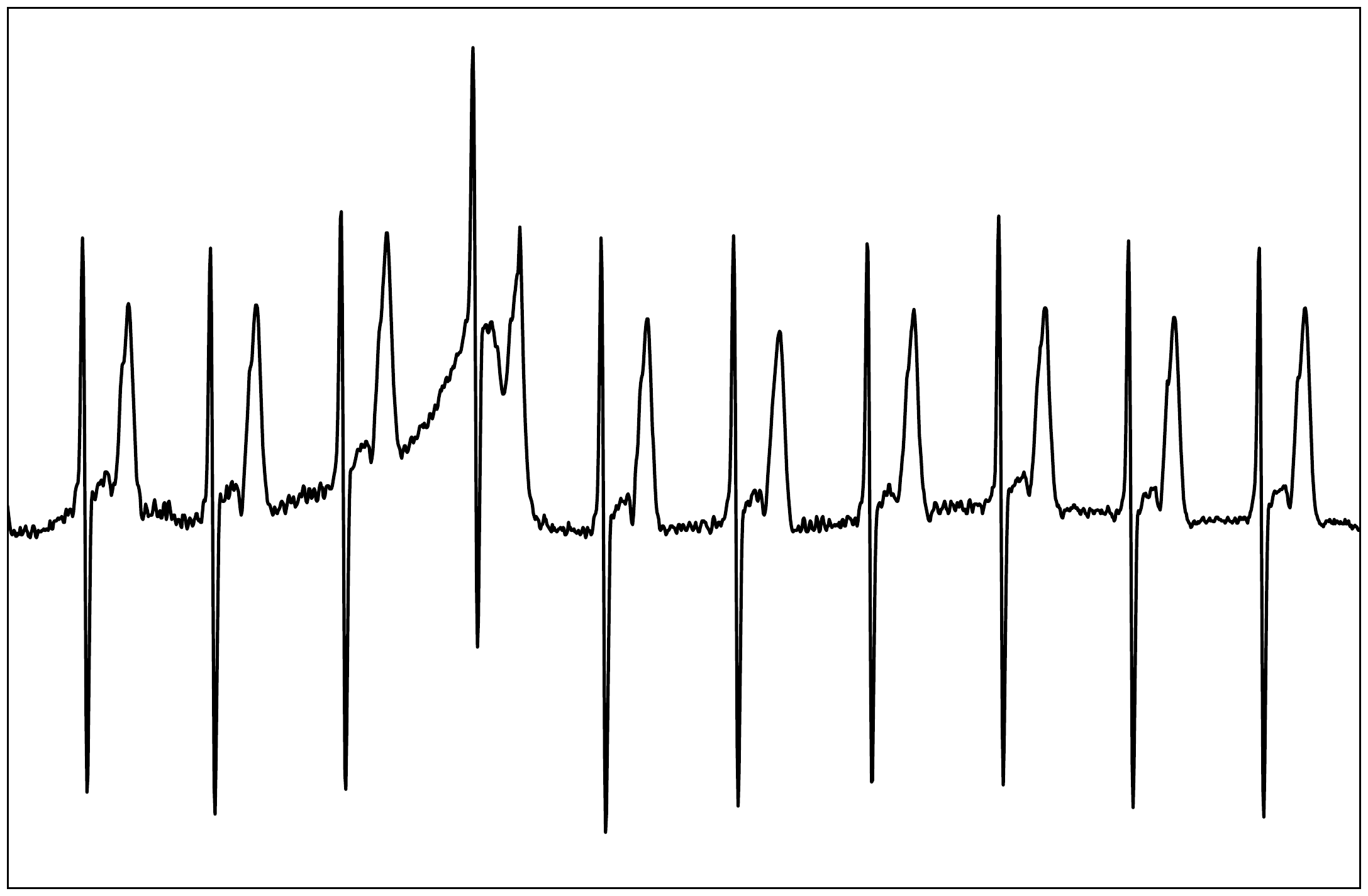}} & 2030 & 36 \\
    NOISE & Noise & \parbox[c]{13em}{\includegraphics[height=8em]{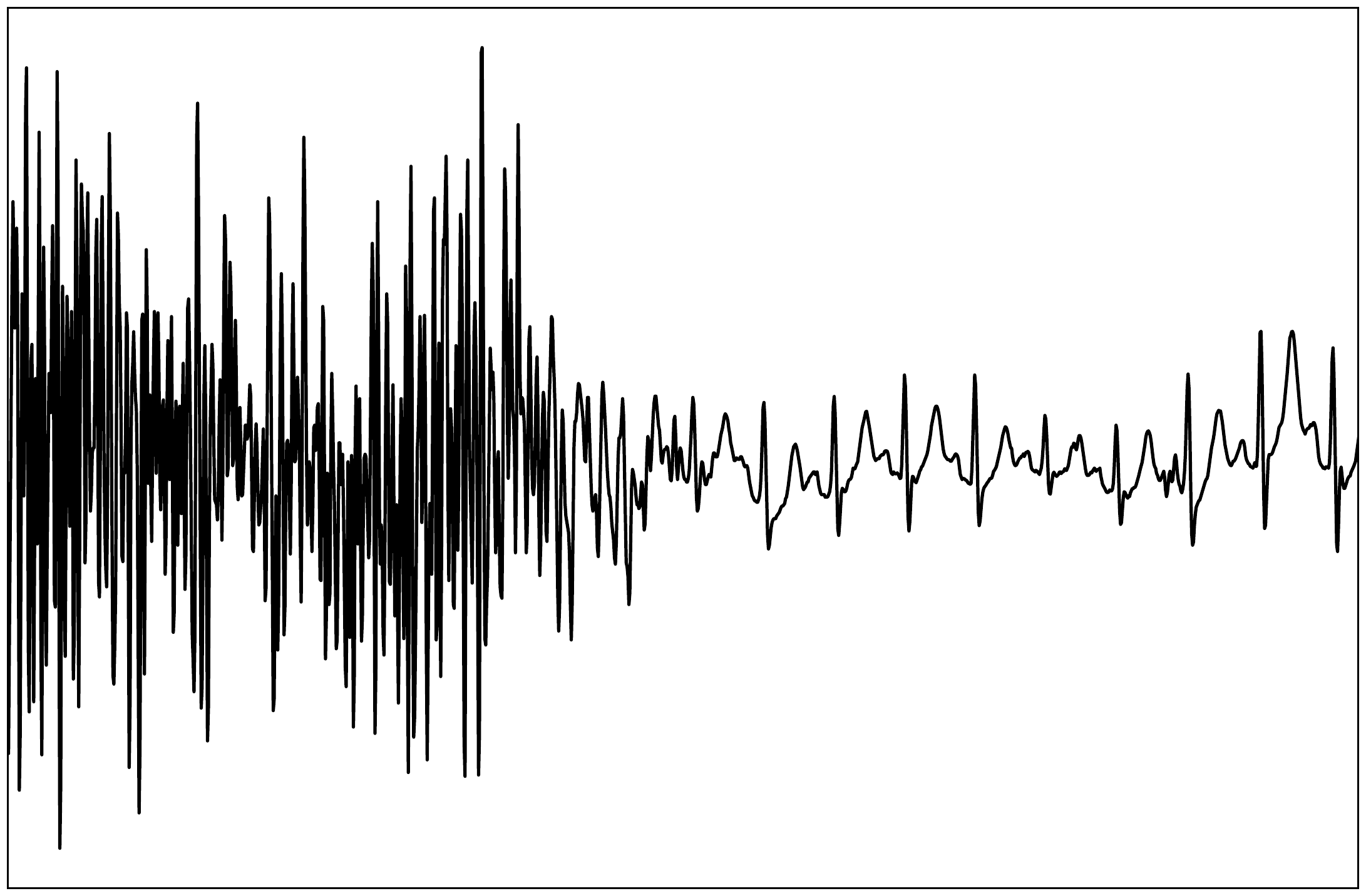}} & 9940 & 41 \\
    SINUS & Sinus Rhythm & \parbox[c]{13em}{\includegraphics[height=8em]{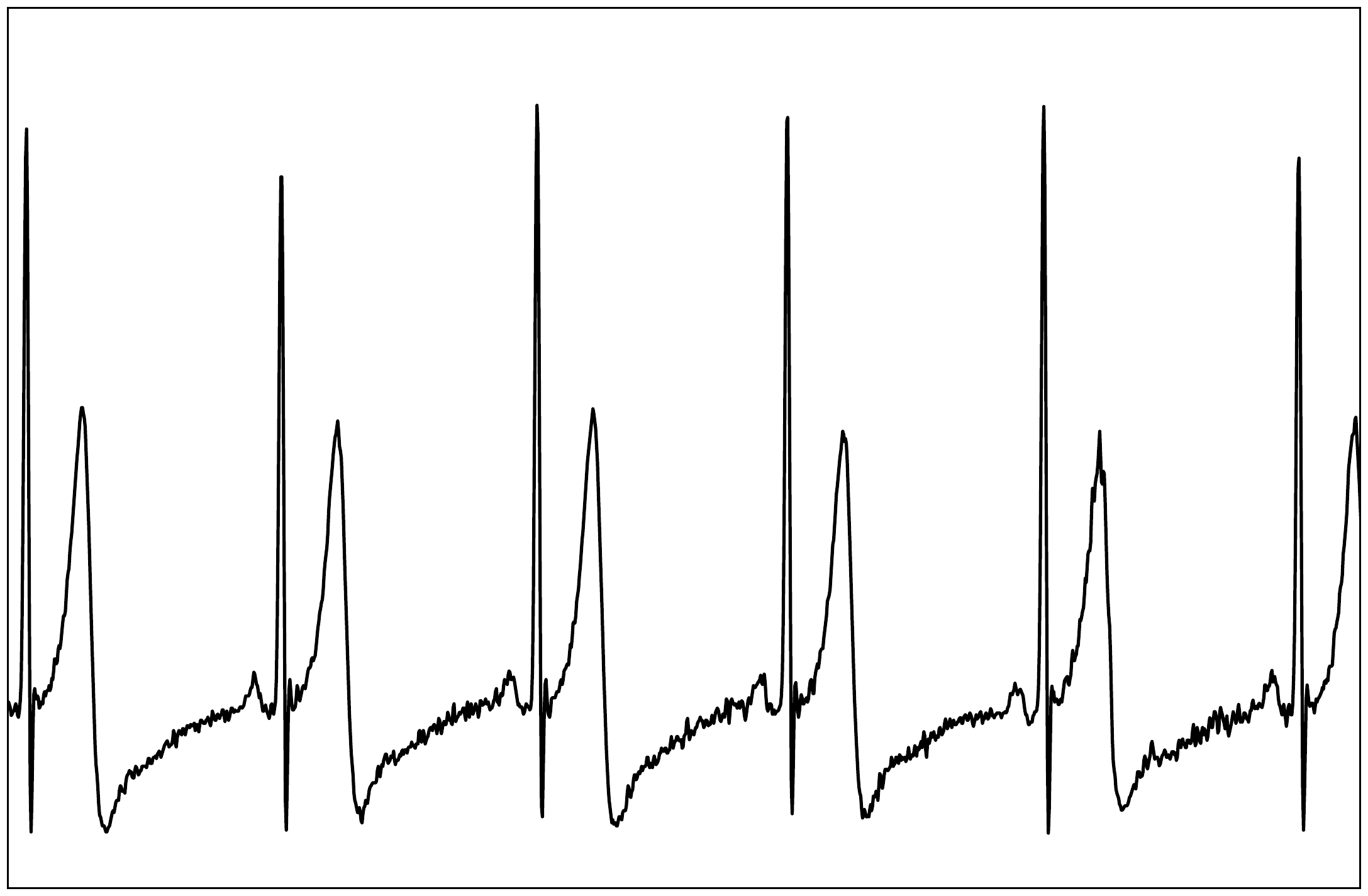}} & 22156  & 215 \\
    SVT & Supraventricular Tachycardia & \parbox[c]{13em}{\includegraphics[height=8em]{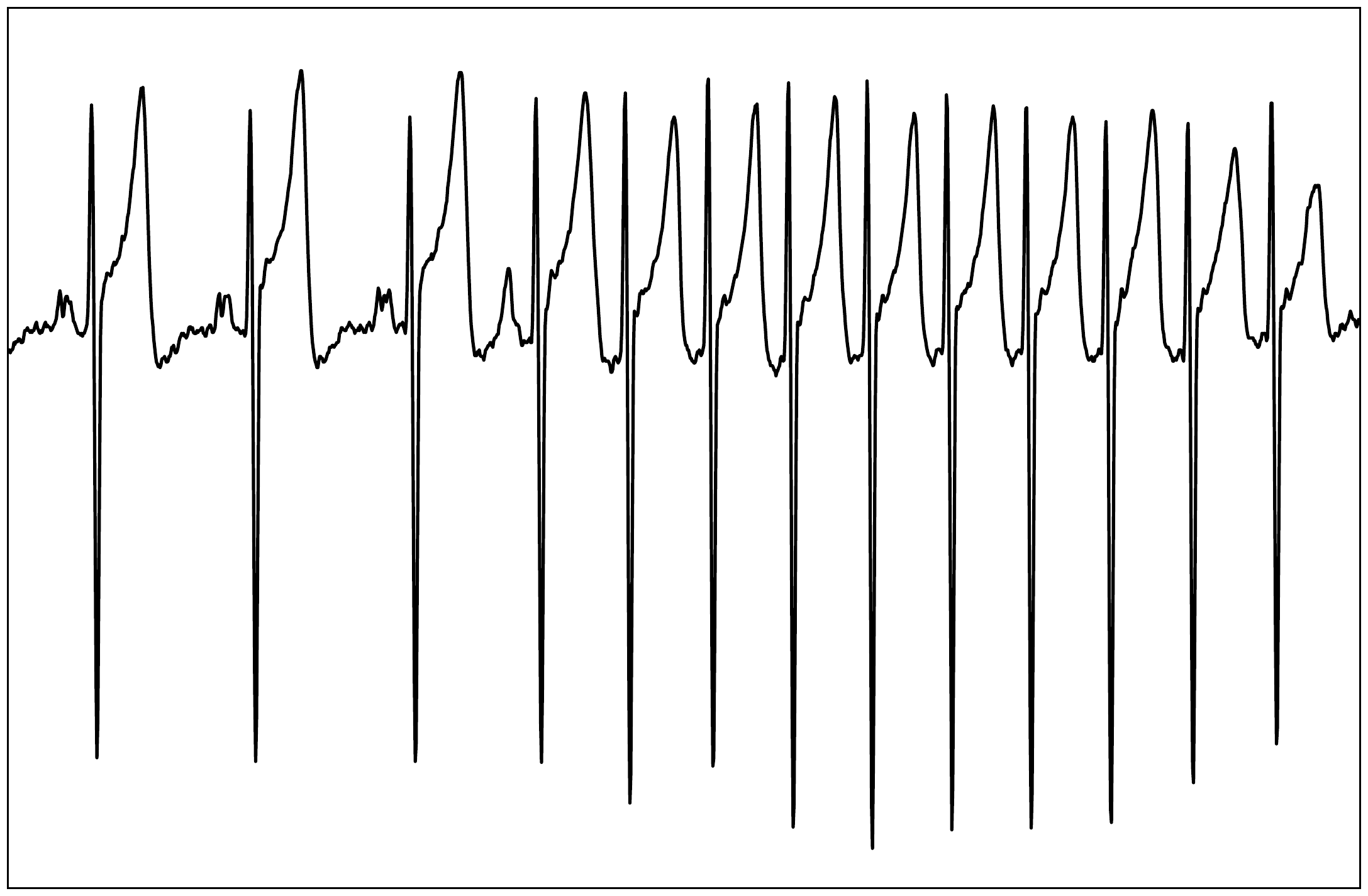}} & 6301 & 34 \\
    TRIGEMINY & Ventricular Trigeminy & \parbox[c]{13em}{\includegraphics[height=8em]{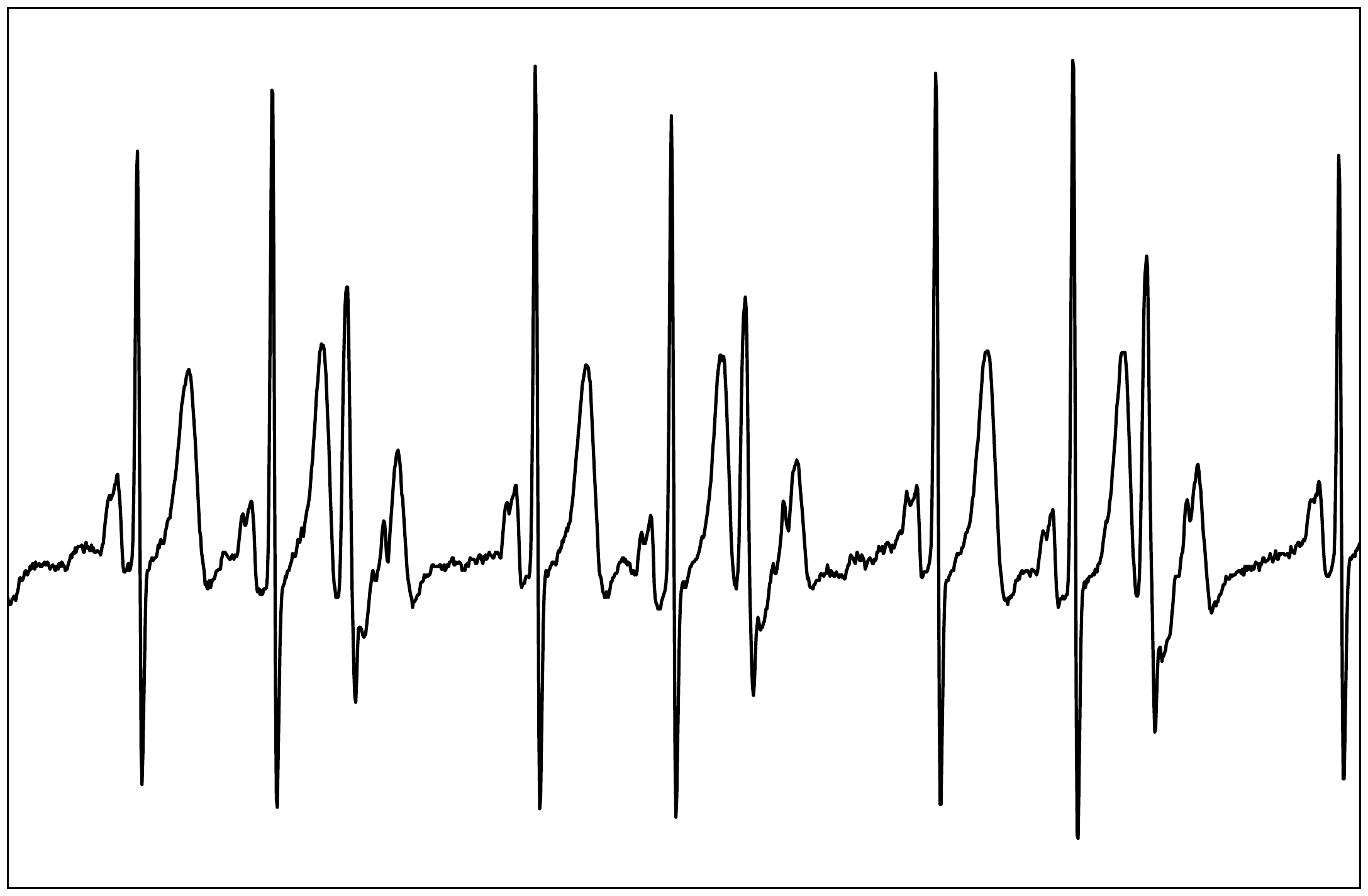}} & 2864 & 21 \\
    VT & Ventricular Tachycardia & \parbox[c]{13em}{\includegraphics[height=8em]{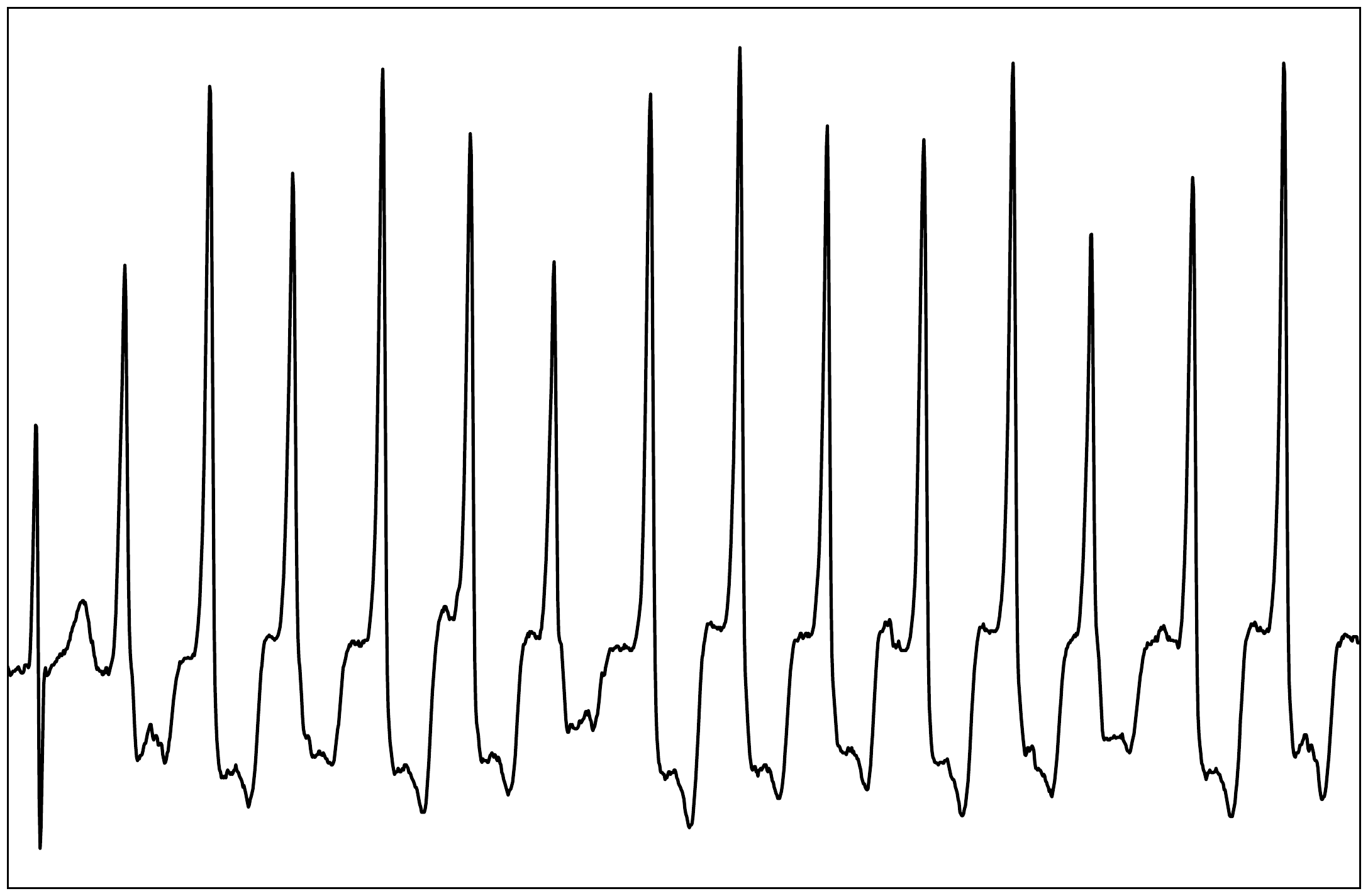}} & 4827 & 17 \\
    WENCKEBACH & Wenckebach (Mobitz I) & \parbox[c]{13em}{\includegraphics[height=8em]{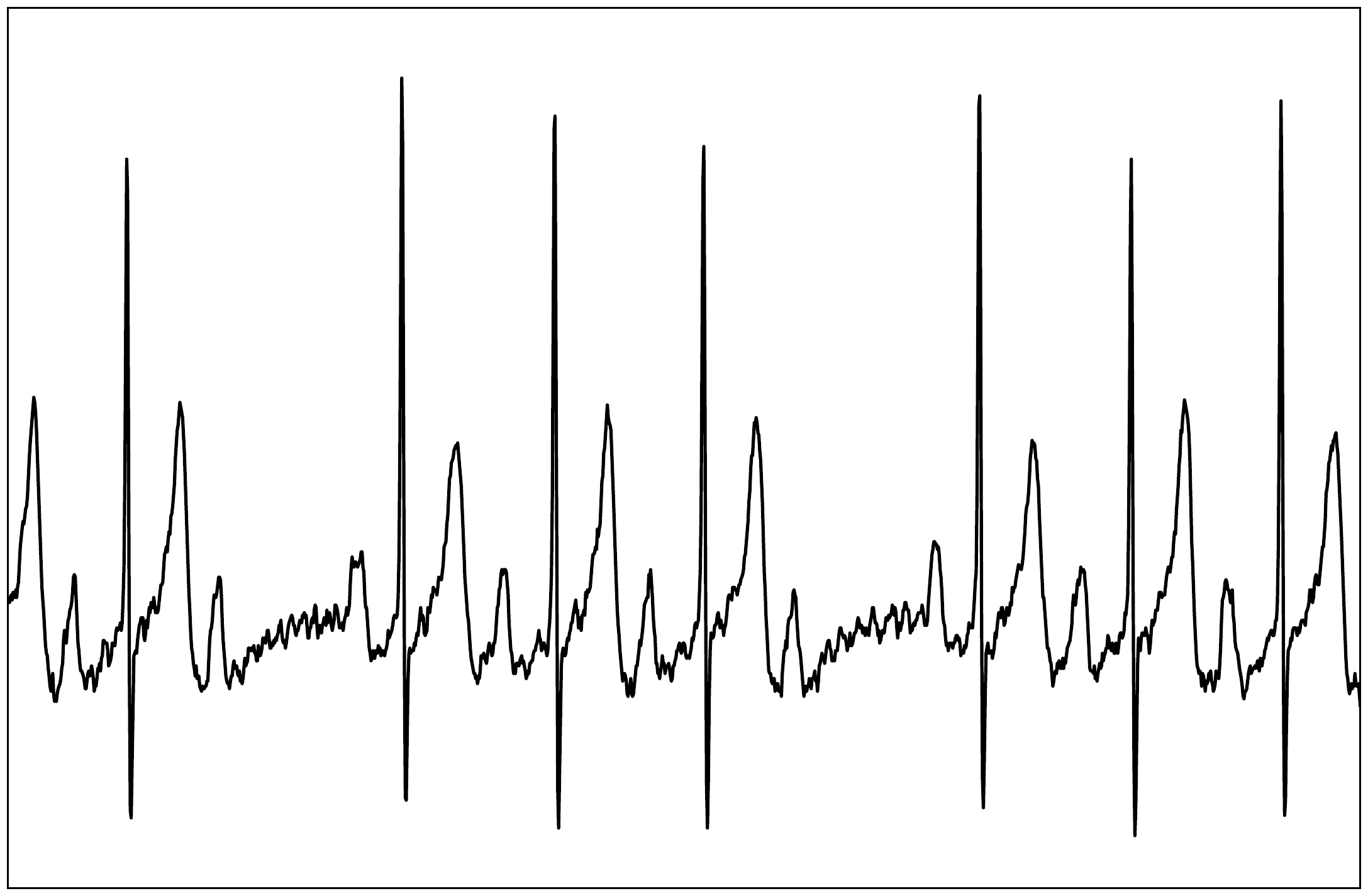}} & 2051 & 29 \\
\end{tabular}
\caption{A list of all of the rhythm types which the model classifies. For each rhythm we give the label name, a more descriptive name and an example chosen from the training set. We also give the total number of patients with each rhythm for both the training and test sets.}
\label{tab:rhythms}
\end{table*}

\end{document}